# A Speech Act Classifier for Persian Texts and its Application in Identifying Rumors

Zoleikha Jahanbakhsh-Nagadeh[1], Mohammad-Reza Feizi-Derakhshi [2*] and Arash Sharifi[3]

1- Department of Computer Engineering, Science and Research Branch, Islamic Azad University, Tehran, Iran.
2*- Department of Computer Engineering University of Tabriz, Tabriz, Iran.
3- Department of Computer Engineering, Science and Research Branch, Islamic Azad University, Tehran, Iran.
[1]zoleikha.jahanbakhsh@srbiau.ac.ir, [2*]mfeizi@tabrizu.ac.ir, and [3]a.sharifi@srbiau.ac.ir

Corresponding author address: Mohammad-Reza Feizi-Derakhshi, Department of Computer Engineering University of Tabriz, Tabriz, Iran.

**Abstract-** Speech Acts (SAs) are one of the important areas of pragmatics, which give us a better understanding of the state of mind of the people and convey an intended language function. Knowledge of the SA of a text can be helpful in analyzing that text in natural language processing applications. This study presents a dictionary-based statistical technique for Persian SA recognition. The proposed technique classifies a text into seven classes of SA based on four criteria: lexical, syntactic, semantic, and surface features. Also, WordNet ontology is utilized to enrich the features dictionary by extracting the synonyms of each word in the input text. To evaluate the proposed technique, we utilized four classification methods including Random Forest (RF), Support Vector Machine (SVM), Naive Bayes (NB), and K-Nearest Neighbors (KNN). The experimental results demonstrated that the proposed method using RF and SVM as the best classifiers achieved a state-of-the-art performance with an accuracy of 0.95 for classification of Persian SAs. Our original vision of this work is introducing an application of SA recognition on social media content, especially identifying the common SA in rumors and its application in the rumor detection. Therefore, the proposed method utilized to determine the common SAs in rumors. The results showed that Persian rumors are often expressed in three SA classes including narrative, question, and threat, and in some cases with the request SA. Also, the evaluation results indicate that SA as a distinctive feature between rumors and non-rumors improves the accuracy of rumor identification from 0.762 (based on common context features) to 0.791 (the combination of common context features and four SA classes).

**Keywords**- Speech Act, Persian text classification, Feature extraction, WordNet, Rumor detection.

## I. INTRODUCTION

Speech Act (SA) is the performed action by a speaker with an utterance. The theory of SA was first Proposed by Austin [2] and refined by Searle [3]. Searle [4] has introduced five categories of SAs: Assertive (e.g. reciting a creed), Directives (e.g. requests, commands, and advice), Commissives (e.g. promises and oaths), Expressives (e.g. congratulations, excuses, and thanks) and Declarations (e.g. baptisms or pronouncing someone husband and wife). Assertive sentences commit the speaker to something being the case and speaker's purpose for transferring the information to hearer; directives cause the hearer to take a particular action and show the speaker's intention; declarative sentences express the speaker's attitudes and emotions. Searle' classification of SAs is known as basic SA taxonomy that is used in many research.

For example, when you say "I'll be there at six", you are not just speaking, but you seem to be performing the SA of "promising". SAs as functional units play an important role in effective communication. We use SAs in our conversations every day when greeting, compliment, request, invitation, apology, threaten, and so on. Normally, the SA is a sentence, but it can be a word like "Sorry!" to perform an apology, or a phrase as long as it follows the rules necessary to accomplish the intention like "I'm sorry for my behavior."[1]. Understanding the SAs of a text can help to improve the analysis of texts and give us a better understanding of the state of mind of the people. The automatic recognition of SA (also known as dialogue act) is known as a necessary work in many Natural Language Processing (NLP) systems such as sentiment classification, topic modeling, and Text-To-Speech (TTS) systems and assertion tracking. In this study, the other application of SA classification is represented to identify the common SAs of rumors. The identification of common SA in rumors can be an important step in recognizing rumors. The main contributions of this study are summarized as follow:

*1) Use WordNet ontology to enrich the Dictionary of Features.*

Extracted features are a valuable set of lexical, semantic, syntactic, and surface features in seven SA classes. These features set have a notable impact on performance and thereby increase the accuracy of SAs classifier. To overcome the limitations of words within the dictionary of each SA class, WordNet ontology is used to find synonyms of words that are not in the dictionary.

*2) Identify common SAs of rumors and its application in rumor detection.*

Rumors are expressed by specific SAs to increase the audience's motivation for rumor distribution. Hence, by analysis of the content of rumor and retrieving the type of their SA and determining the common SA in rumors, a preliminary step can be taken to identify rumors.

The rest of the paper is organized as follows: Section 2 discusses related works that have done on SA classification. In Section 3 are introduced applied datasets in this work. Section 4 describes the proposed methodology for identifying the SAs on the Persian language. Then in section 5, the application of SA classifier in determining the common SAs of rumors is discussed, and in Section 6 results of experiments and evaluations and conclusions of paper is shown. In section 7, we conclude with a discussion and directions for future work.

## II. RELATED WORKS

The problem of identifying the SA has been a field of interest to researchers from several areas for a long time. The SA classification has been studied not only in English but also in many other languages such as Chinese [25], Korean [26], Arabic [11] and so on. However, in the Persian language is done very little work on automatic SA. As far as we know, the only published work on Persian SA classification is the work of Soltani-Panah et al. [1]. In following is mentioned some research related to our work:

In 1999, Klaus Ries [19] presented an incremental lattice generation approach to detect SA for spontaneous and overlapping speech in telephone conversations (CallHome Spanish). He used a Hidden Markov Models (HMM) algorithm where the states are SAs and the symbols emitted are sentences, and also used neural network based estimates for the output distributions. He showed the performance of neural networks in the SA classification.

Vosoughi et al. [5] proposed SA recognition on Twitter (English tweets). Vosoughi created a taxonomy of six SAs for Twitter and proposed the set of semantic and syntactic features for labeling tweets manually. Then, the labeled dataset is used to train a logistic regression classifier. In another two work on the English language is done by Zhang et al. for SA classification on Twitter by using supervised [6] and semi-supervised [7] methods. They proposed the set of word-based features and character-based features for creating a labeled dataset.

Qadir and Riloff [18] performed the SA classification on a collection of message board posts in the domain of veterinary medicine. They create a sentence classifier to recognize sentences containing four different SA classes: Commissives, Directives, Expressives, and Representatives. Qadir and Riloff used the collection of features, including lexical and syntactic features, SA word lists from external resources, and domain-specific semantic class features.

Sherkawi et al. [11] presented rule-based and statistical-based techniques for Arabic SA recognition. They represent that advantage of building an expert system is that it did not require a large corpus; instead, from a small set of examples, a core expert system is built. In contrast, using machine-learning methods is a more time-saving task, but it requires a large corpus for training. They evaluated the proposed techniques using surface features, cue words, and context features; and found that the cue words feature set is simple and indicative of the SAs when using machine-learning methods. In contrast, the expert system performance has improved significantly when adding context features.

Král et al. [21] used syntactic features derived from a deep parse tree to recognize dialogue acts in the Czech language based on conditional random fields. They considered two types of features: baseline and syntactic. The baseline features are: words inflected form, lemmas, part-of-speech tags, pronoun or adverb at the beginning of the sentence, and verb at the beginning of the sentence. The syntactic features rely on a dependency parse tree: dependency label, root position in the utterance, unexpressed subjects, and basic composite pair (subject-verb inversion).

In 2010, Soltani-Panah et al. [1] presented the first work on the automatic categorization of Persian texts based on speech act. They considered seven classes of SAs in Persian language texts and three classification methods including Naïve Bayes (NB), K-Nearest Neighbors (KNN) and Tree learner. Soltani-Panah et al. evaluated three classification algorithm on Persian dataset. They concluded that the KNN with an accuracy of 72% is the most efficient algorithm for classifying Persian SAs.

Our work is similar to work [5], [6], [18] [11] but what is notable in our work relative to previous works is that we applied WordNet ontology to enrich the dictionary of the features of each SA class by extracting synonyms of words in the input text.

## III. DATA

In this work, the supervised classifiers are utilized to classify SAs, so like any other supervised classification problem, a large labeled dataset is needed. Also, to analyze and detect the common SA in Persian rumors, it is necessary to provide a collection of Persian rumor texts. Due to the lack of such data, we manually collected and labeled this dataset in two classes include, rumor and non-rumor.

### A. SA dataset

To train supervised algorithms and evaluate our proposed SA classifier, we employed the labeled dataset by Soltani-Panah et al. [1] from multiple subjects and sources. They labeled the raw corpus that is created by the Research Center of Intelligent Signal Processing (RCISP).

TABLE I
SIX SA TYPES ALONG WITH DESCRIPTIONS

| SAs types (Abbr.) | Description |
|---|---|
| Questions (Ques) | These are usual questions for information or confirmation. |
| Requests (Req) | Politely asks from somebody to do or stop doing something. |
| Directives (Dir) | With these SAs we cause the hearer to take a particular action perforce. |
| Threats (Thrt) | With these SAs we can promise for hurting somebody or doing something if hearer does not do what we want. |
| Quotations (Quot) | These are SAs that another person said or wrote before. |
| Declarative (Declar) | Transfer information to hearer, these commit the speaker to something being the case. |
| Narratives (Narrov) | These SAs report connected events, real or imaginary. With these SAs we tell what has happened |

This dataset contains labeled 9145 Persian sentences in seven SA categories. The SAs are compiled into a database of 1734 Questions (Ques), 928 Requests (Req), 1113 Directives (Dir), 544 Threats (Thrt), 850 Quotations (Quot), 2000 Declaratives (Declar), and 1976 Narratives (Narrov). The texts of this training dataset are gathered from different sources such as newspapers, magazines, journals, internet, books, weblogs, itineraries, diaries, and letters. This data includes various domains such as Economy, Export, Culture, Sciences, etc. These six SA types along with descriptions are listed in Table I.

### B. Rumors and non-rumors dataset

In this study, we intend to detect the common SA in rumors in the Persian language. Thereby, a dataset of a few thousand Persian posts of Telegram channels in Iran from May 1, 2017, to March 30, 2018, is collected. For this purpose, we utilized the provided API by Computerized Intelligent Systems (ComInSys)[1] of the University of Tabriz [15]. Using this API, we crawled the rumors of three public Telegram channels of three Iranian websites, Gomaneh.com, wikihoax.org and Shayeaat.ir which review Persian rumors. Also, several public Telegram channels are reviewed to collect news released on the Telegram, including Fars News Agency, Iranian Students' News Agency (ISNA), Tasnim News Agency, Tabnak, Islamic Republic News Agency (IRNA). Thereby, 882 rumors and 1093 non-rumors is collected.

### C. Features

The effectiveness of speech act classification task depends basically on the features used in training the classification model [11]. We collected a set of 2275 content features as useful features that can distinguish six SA classes in the Persian language. These features can be divided into four categories: Lexical, Semantic, Syntactic and Surface.

#### 1) Lexical Features

- **Particular words.** Particular words give us valuable information about the type of SAs. So, we collected the particular words in the 6 categories based on SAs categories that these words often appear in texts with specific SAs. To generate this set, we extract N-grams from sequences of the training dataset.

- **Cue words.** There are words that are explicit indicators of the SA also key to understanding, such as, the question mark is a base feature for *Ques* sentences and next base feature is question words (such as "چه"/ce/what, "چطور"/cetowr/how, "چرا"/cerâ/why). The base feature of *Req* sentences is conditional words (such as "لطفا"/lotfan/Please, "اگه ممکنه"/age momken/ if possible).

#### 2) Semantic Features

- **N-grams.** We extract N-grams unigram, bigram phrases from the text. A unigram is a one-word sequence of words like ("برو"/boro/Go). A 2-gram (or bigram) is a two-word sequence of words like ("لطفا برو"/lotfan boro/Please go.

- **Vulgar words.** Vulgar words contain the set of slang and obscene (bad) words. We collect a total of 965 Persian vulgar words from an online collection of vulgar words. Vulgar words mostly appear in the threats SA class. This feature has a binary value that indicates whether vulgar words appear in the text or do not appear.

- **Speech act verbs.** SA verbs are content words which can be used to describe types of SAs. In other words, SA verb is a verb that explicitly conveys the kind of SA being performed, such as promise, apologize, predict, request, warn, insist, and forbid. Also, it is known as the performative verb. Based on SA verbs in English, we collected 910 SA verbs for Persian in 6 classes.

- **Sentiment.** We believe that sentimental polarity of a text could be an informative factor to identify SAs. For example, texts with *Thrt* SA are usually dominated by negative sentiment. To calculate the sentiment score of the text, we utilized a lexicon-based method that the sentiment polarity of Persian words is obtained using the NRC Emotion Lexicon. NRC is created by Saif et. al. [21] at the National Research Council Canada, which consists of 14183 words that each word is tagged with one of the sentiment labels; Positive, Negative, or Neutral. So the sentiment score of the text is calculated based on the polarity of the words using equation 1.

---

[1] www.cominsys.ir

$$Score_{Sent}(T) = \frac{|PSntm(T)| - |NSntm(T)|}{|PSntm(T)| + |NSntm(T)|} \quad (1)$$

Where, $PSntm$ and $NSntm$ respectively are the number of words with positive and negative polarity in the text T.

3) *Syntactic Features*

- **Part-of-Speech (POS) tags.** POS is syntactic categories for words. We applied an HMM-based POS tagger for Persian POS tagging. Interjections and words with the "IF" tag are mostly used in the requested sentences. Also, adjectives can appear in Declarative and Narrative texts.
- **Punctuations.** We consider three punctuations: '?', '!', and ':'. Punctuations can be predictive of the SA in a sentence. For example, the punctuation '?' appears in *Ques* or *Req* sentences, while punctuation '!' appears in "Dir", *Req*, or *Thre* sentences, and punctuation ':' appears in the "quotation" sentences. These punctuations are binary features that this binary value indicates to presence or absence of these symbols. Previous work such as Mendoza et al. [27] has shown that only one third of tweets with question marks are real questions, and not all questions are related to rumors. Also, in Persian language, the question mark is not specific to *question* sentences, but rather often appearing in *request* sentences and, in limited cases, in *threat* sentences.

4) *Surface features*

- **Token position.** We have defined an assumption similar to the assumption of Moldovan et al. [22]. Based on this assumption, the first and last words in a sentence can be valuable indicators in determining the SA of texts. One argument in favor of this assumption is the evidence that hearers start responding immediately (within milliseconds) or sometimes before speakers finish their utterances [23].

### IV. PROPOSED METHODOLOGY

In this study, we used supervised algorithms to categorize texts based on speech acts. The effectiveness of speech act classification task depends basically on the features and a large labeled dataset used in training the classification model [11]. So we focused on the contextual features that can give valuable information about the SA of the text to the classifier. These features are used to create feature vectors with informative elements and lower dimensional.

Since each word may contain several synonyms, and if the original word is not in the dictionary but its synonyms are in the dictionary, that word is not considered as a feature because it is not found in the dictionary. Therefore, the features in the dictionary of Words are not enough for text analysis. To address this problem, we used WordNet ontology to find synonyms for any words in the text that were not found in the dictionary. So the feature vectors are created with more accurate information. Then, the classifier is fed by these vectors, which leads to the creation of a robust system to identify the SAs of Persian texts. Below, each of the steps of the proposed method is explained in detail. The general structure of the proposed method for SA classification is shown in Figure 1.

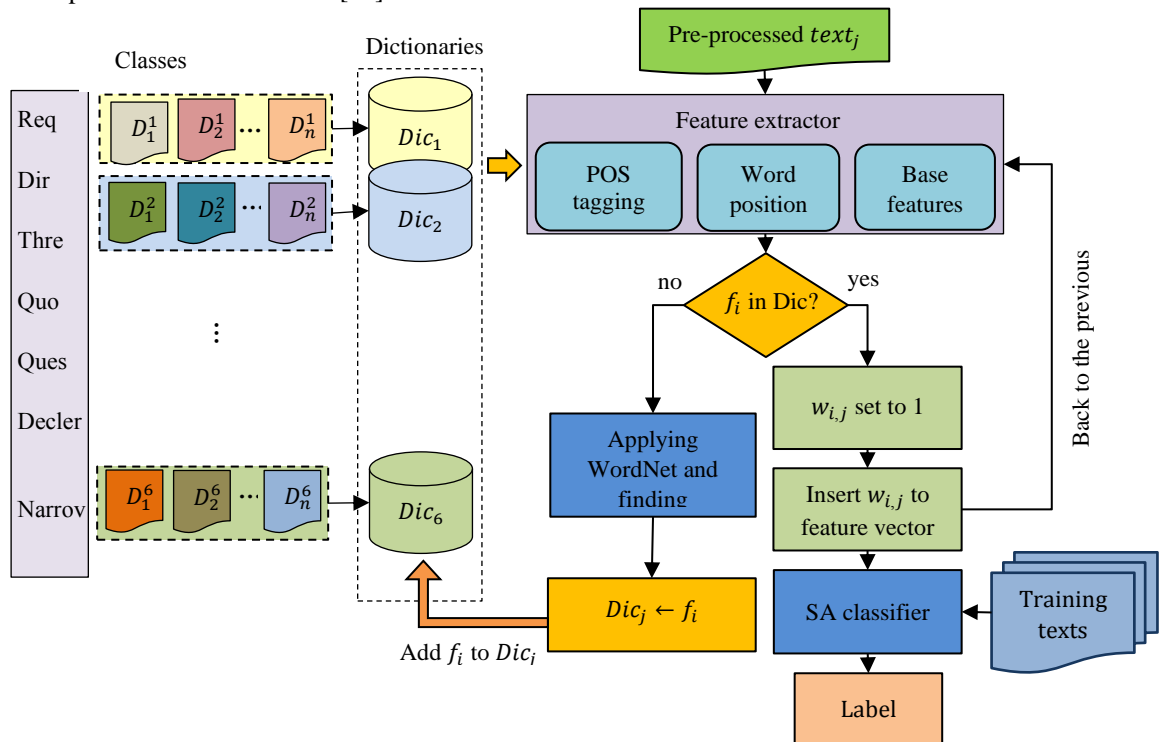

**Fig. 1. Process of creating dictionary of features for each SA class and SA classification**.

*A. Text Pre-processing*

At this stage, useless information of the text is eliminated. Pre-processing reduces the noise in text and hence makes the data more 'clean' and capable of training the classifier more effectively. In this study, we use the four steps of preprocessing include: tokenization, removing stop words, normalization, stemming and lemmatization.

*1) Normalization*
The process of transforming text into a single canonical form that it might not have had before.

*2) Tokenization*
This is the process of splitting a text into individual words (unigram) or sequences of words (bigram).

*3) Stemming word*
Reducing inflectional forms and sometimes derivationally related forms of a word to a base form.

*4) Lemmatization*
Reducing various linguistic forms of a word to their common canonical form.

*B. Feature Extraction*

In text classification, a major problem is the high dimensionality of the feature space. Therefore, using feature extraction methods, a set of most informative and indicative features of the training set are extracted [8]. In this study, four different methods have been used to extract features. These methods are described following.

*1) Synonym Extractor*
Synonym extraction can be useful in feature dictionary enrichment and more accurate feature extraction. So, we utilized FarsNet as a lexical ontology for the Persian language to extract the synonyms of each word within text. FarsNet is developed by the National Language Processing Laboratory as the first Persian WordNet [20].

*2) HMM-based Parts-of-Speech Tagging*
Hidden Markov Models are suitable for POS tagging in Persian language and quite comparable with best-case results reported from other models. Therefore, we used Orumchian tagger [12] for Persian Parts-of-Speech (POS) tagging. Orumchian tagging system follows the TNT POS tagger [13]. The TNT tagger is based on HMM theory. The tagging system uses 2.5 million of tagged words (Bijankhan corpus) as training data and tag-set contains 38 tags. The obtained accuracy of this approach is 96.64%. In this study, five tags of noun, six tags of adjective, six tags of adverb, six tags of verbs and 11 tags include punctuation, connection phrases and so on are selected to analyze the text.

*3) Feature Extraction Based on Word Position*
Each word in the text has a specific location that it can be defined as word position. Useful information can be extracted from different word positions in the text. For example, in the English language, the first few words of a dialog utterance are very informative of that utterances speech act [22]. This principle applies in the Persian language too. For example, in the Persian language, Req words usually appear at the beginning of the Req sentences, such as "لطفا در را باز کن" /"Please open the door.", and in some cases appear at the end of the sentence, such as "در را باز کن لطفا" /"Open the door, please.". The word "please" is a "requested" word and its position in a sentence is mostly at the beginning or end. Therefore, in each text, the first and end words are extracted. If any of these words are found in a dictionary (common words) of one of the SA classes, its binary property is set to 1. Also, in Ques sentences, question words appear at the beginning of the sentence.

In another sentence such as " من از دوستم درخواست کردم تا کتابش را به من بدهد". /I asked my friend to give me his book", although the word "ask" (as an SA verb of Req class) appears in the sentence, but the sentence's SA is not Req. In such cases, the word position in the text can be helpful, that is, a word is considered as a base word for SA of Req class when that word appears at the beginning or at the end of the sentence. Thus, the "word position" as a binary feature indicating whether these words appear in the first position or end of the text.

*4) Extraction of Base Features.*
In some cases, the identification of an SA is really difficult. For example, to evaluate the category of Ques SA, consider the following statements:

I. "آیا این کاغذ را می‌بینی؟" / "âyâ in kâghaz râ mibini?" / Do you see this paper?
II. "نظرتان درباره‌ی تحولات قرن بیستم چیست؟" / nazaretân darbâreie tahavvolâte garne bistom chist? / What is your opinion about the developments of the twentieth century?
III. "ممکنه خواهش کنم به من کمی آب بدی؟" / "momkene khâhesh konam be man kami âb bedi?" / May I ask you give me some water?
IV. "میشه فردا برین؟" / "mishe fardâ berin?" / Is it possible to go tomorrow?
V. "لطفا تلویزیون را روشن کن." / "lotfan telvezion râ roshan kon." / Please turn on the TV.

In the "Question" sentences usually appear the question words along with question marks. The first three sentences (i.e., I and II) are expressed with the Ques SA and three next sentences (i.e., III, IV, and V) with the Req SA. Statement "I" has a question word, but statement II doesn't contain question words. Thus, the SA type of the statement can be identified by the question mark. On the other hand, the sentences "IV" and "V" contain a question mark, but their SA is Req.

To solve this problem, we used two features, including the word position and the base features. For example, in the Req sentences, the first word of the sentence is a Req term, such as "ممکنه"/momkene/May, and the sentence ends with a question mark. In the Persian language, the Ques sentences begin with a question word such as "چرا"/cherâ/why and the sentence ends with a question mark. In question sentences, the first base feature is the question words and the second base feature is the question mark. But Req sentences have only one base feature that is, the Req words. So if the Req words appear at the beginning of the sentence, then the question mark is not considered as the base feature (such as sentence

IV). So we have six binary features for the base features of each of the six SA classes.

*C. Create Enrich Dictionary of Features*

After pre-processing noisy data and extracting valuable features from training dataset, we intend to create an influential classification system by enriching extracted features. To enrich collected features, we used WordNet ontology as a tool. Each Persian word can have several synonyms. WordNet is a lexical database, which groups nouns, verbs, adjectives and adverbs into sets of synonyms, each expressing a distinct concept. We employed WordNet developed in [20] to find synonyms of given Persian words.

In order to, a vector for each text is derived. Vector elements include features that are obtained using the proposed methods for extracting feature. Thus, a subset F′ from F is selected to provide a more efficient description of the documents. To calculate the features weight, we used features frequency in the text.

Classification of text documents involves assigning a text document to a set of pre-defined classes. Since these classes are pre-defined, it is a supervised machine learning task. Therefore, a labeled dataset is needed for training classifier. Let $D = \{D_1^j, D_2^j, D_3^j, \ldots, D_n^j\}$ be the set of $n$ training document of class $C_j: j = 1,2,3,\ldots,k$ (k denotes number of categories, k=6) and let $f_i^j = \{f_{i1}^j, f_{i2}^j, f_{i3}^j, \ldots, f_{im}^j\}$ be the set of $m$ features of the text document $D_i^j$ of the class $C_j$.

We have extracted all features of $C_j$ class as distinctive features of SAs to construct dependent dictionary $Dic_j$ of class $C_j$. Therefore, the dictionary of features for each SA class is collected based on their contribution to discriminate an SA from other SAs. The items of dictionary $Dic_j$ are extracted using three feature extraction methods from training dataset of class $C_j$. But since different writers may use different synonyms of words to express their intention, so it is not enough to use a static dictionary of features to analyze the text and classify it.

To resolve this problem, we used WordNet ontology for the Persian language to get synonyms of a word. For this purpose, each word of the input text that does not find an equation for it in the dictionary of SA classes, it is given to the WordNet. WordNet finds the closest synonyms (Synset) for this Persian word. If related word suggested by the WordNet appears in the dictionary of SA class $C_j$, In this case, the new word will be added to the dictionary $Dic_j$. Then one binary feature indicates that this word has appeared in $Dic_j$. Hence, whenever the system determines the SA of a new text, the dictionary of features is developed with new words that their synonyms are in the dictionary.

## V. SPEECH ACT IN RUMORS

In this section, we will focus on the application of SA classification in identifying the common SAs in rumors. Because, identifying rumors SA can be helpful in the rumors detection. Therefore, first, we need to provide a clear definition of rumors. We define a rumor to an unverified information that is arisen in situations of ambiguity, threat and it is spread among users in a network. According to the research of Computer Emergency Response Team/Coordination Center (CERTCC)[2], five common types of rumors in cyberspace in Iran are as follows:

1) *False news of famous figures*
   Perhaps one of the most common rumors in the virtual world is the false news of the death of famous persons and well-known figures. This phenomenon has become widespread in the western countries, and unfortunately, in recent years there has been an increasing trend in Iran, such as news of the deaths of artists, actors, footballers, political figures and so on.
2) *Fake messages with emotional content*
   These types of rumors have been common in e-mail services, and have recently been published on social networks. Generally, these messages include content such as helping find a missing child with a request to retype a story that they are trying to spread this fake news by stimulating the audience's emotions.
3) *Rumors of electoral*
   Election issues are one of the most important and most sensitive political issues in society. Because of the importance and sensitivity of the subject and the public's attention to it, news related to the election is spreading at a great rate, especially on social networks, and stimulates political currents and individuals.
4) *False news about social networks*
   One of the most common rumors on social networks is the creation of suspicion about social networks.
5) *Rumors Related to risks*
   This can be considered one of the most influential social phenomena derived from social networks. This phenomenon causes widespread social anxiety and disturbs public opinion and even causes widespread insecurity and distrust to state institutions.

The SA of rumors in first and third types are usually declarative or narrative. The second type of rumors is often expressed by the request SA. The third type of rumors are often expressed by declarative SA, and rumors of the fourth type are often expressed by question SA. The fifth type of rumors are expressed by threat SA.

Based on this evaluation, it is clear that people who create rumors try to persuade people to accept their rumors. To this end, they incite fear and anxiety in the audience. Therefore, these people use narrative, question, threat, and request SAs to achieve their goals. Determining the SA of rumors can play a significant role in the auto-rumor detection system. In the next section, the results of the experiments show that rumors are often expressed by what type of SAs.

## VI. EXPERIMENTS AND DISCUSSION

To evaluate the performance of the proposed method for

---

[2] https://www.certcc.ir/

SA recognition, we ran experiments on seven sets of the labeled dataset by using four different supervised classifiers include: Random Forest (RF), Support Vector Machine (SVM), Naive Bayes (NB), and K-Nearest Neighbors (KNN). We used 10-fold cross-validation methods for training and testing classifiers. Also, to evaluate our system, we used the performance metrics such as Precision, Recall, and F-measure.

TABLE II
COMPARISON OF FOUR CLASSIFIER FOR SPEECH ACT CLASSIFICATION USING FE-SA METHOD.

|     | Ques  | Req   | Dir   | Thrt  | Quot  | Dec   | Narrv | Avg   |
|-----|-------|-------|-------|-------|-------|-------|-------|-------|
| RF  | 0.93  | 0.917 | 0.918 | 0.891 | 0.73  | 0.996 | 0.941 | **0.903** |
| SVM | 0.931 | 0.917 | 0.916 | 0.894 | 0.728 | 0.996 | 0.942 | **0.903** |
| NB  | 0.916 | 0.916 | 0.905 | 0.878 | 0.749 | 0.997 | 0.93  | 0.899 |
| KNN | 0.928 | 0.915 | 0.916 | 0.892 | 0.728 | 0.997 | 0.931 | 0.901 |

TABLE III
COMPARISON OF FOUR CLASSIFIER FOR SPEECH ACT CLASSIFICATION USING FE-WN-SA METHOD.

|     | Ques  | Req   | Dir   | Thrt  | Quot  | Dec   | Narrv | Avg   |
|-----|-------|-------|-------|-------|-------|-------|-------|-------|
| RF  | 0.957 | 0.957 | 0.919 | 0.985 | 0.851 | 1     | 0.971 | **0.949** |
| SVM | 0.949 | 0.955 | 0.917 | 0.982 | 0.842 | 0.996 | 0.971 | **0.945** |
| NB  | 0.931 | 0.951 | 0.912 | 0.951 | 0.827 | 0.997 | 0.962 | 0.933 |
| KNN | 0.947 | 0.951 | 0.918 | 0.967 | 0.801 | 0.997 | 0.967 | 0.935 |

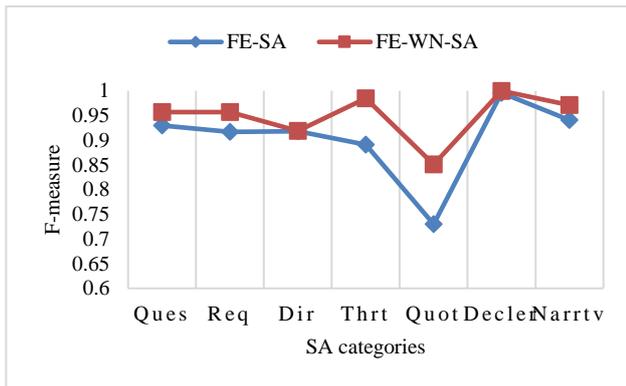

**Fig 2. Comparison of F-measure of FE-SA and FE-WN-SA in SA classification.**

*A. Impact of WordNet Ontology in Classification*

The proposed techniques are performed on SA classification two times; the first time, using FE techniques but without using WordNet ontology (that called FE-SA) and the second time using FE techniques and WordNet ontology (that called FE-WN-SA) by four classifier RF, SVM, NB, and KNN. The evaluation results of these two experiments are presented in Tables II and III, respectively. The results of tables II and III proved that the FE-WN-SA method more effective than FE-SA method in speech act classification. Therefore, it should be noted that the using of WordNet as a tool for extracting synonyms can have a significant effect on the classification improvement.

Based on best experimental results of FE-WN-SA (Table III), the accuracy of RF and SVM as the best classifiers were 0.95, NB as a fast classifier showed a performance of 0.93, the accuracy of KNN as the slow classifier was 0.94. The experimental results of the proposed method for enriching dictionary of features for each SA class and classifying Persian texts based on SA show a great improvement. The accuracy of the proposed method based on classifiers RF and SVM is 0.95. As it can be understood from Fig 2., FE-WN-SA method generally has better performance than FE-SA, except in Dir class; the reason behind this is that the WordNet does not contain imperative verbs inherently.

*B. Comparison*

Since, in this study are utilized a set of lexical features such as Particular words, Cue words, SA verbs on Persian language to classify texts based on SA, so proposed method is specific to Persian language. Thereby, the performance of our proposed SA classifier is compared to the only done work on Persian text SA classification that it is proposed by Soltani-Panah et al. [1]. This comparison is performed on a same dataset of Persian speech acts that compiled and labeled by Soltani Panah et al. [1]. Table IV shows the results accuracy of our proposed method compared to [1]. Our classifier outperformed the Soltani-Panah classifiers. Soltani-Panah has encoded all the words in a sentence as an element of each feature vector except functional words such as conjunctions, propositions, numbers, and surnames, etc. But, we utilized set useful features as distinctive characteristics for seven SA classes. Also, by extracting synonym words using WordNet, we were able to develop a set of features in the dictionary of SA classes. As with the enrichment of the dictionary of features, the system can detect the speech act of new texts.

In order to, the F_measure score of classification from 74% (Soltani-Panah's result) to 95% (our result) is increased. This result demonstrates that the use of FE techniques and WordNet ontology for extending the dictionary of common words in each SA class can be effective in improving the efficiency of the classifier.

TABLE IV
RESULTS ACCURACY OF OUR PROPOSED METHOD COMPARED TO THE SOLTANI-PANAH CLASSIFIERS.

| Method | Soltani-Panah | Proposed method |
|--------|---------------|-----------------|
| RF     | -             | **0.949**       |
| SVM    | -             | 0.945           |
| NB     | 0.739         | 0.933           |
| KNN    | 0.72          | 0.935           |

*C. Statistical analysis*

In this study, we intend to apply the proposed SA classifier to identify the common SAs in Persian rumors. To this end, we collected a set of Telegram posts. Of a few thousand collected Telegram posts, 882 rumor and 1093 non-rumors are collected. Since rumors spread in the various fields, such as political, economic, sports, and so on, we requested our annotators to manually label the collected rumors in six categories of political, economic, events, sports, cultural, and health and medicine. In the labeled dataset, events news with 205 rumors (23%) has the largest share. Political news with 186 rumors (21%) is in the next place. Within this timeframe, 154 economic news (17%) and 128 sports news (15%) have been denied. Health and medicine news is at the next place with 114 rumors (13%). Finally, cultural news with 95 rumors (11%) has the lowest share. The classification of rumors in these six categories is due to the fact that the SAs of political rumors can be different from the SAs of cultural rumors or other rumors. To identify the common SAs in rumors, we first need to examine rumors in different domains.

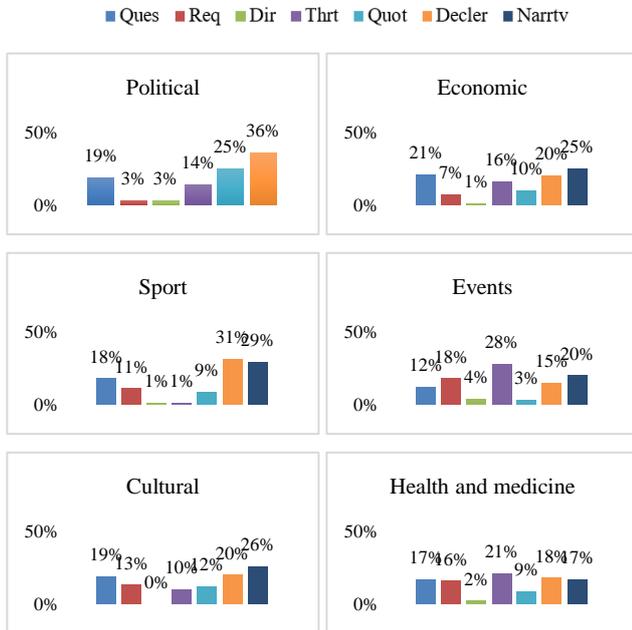

**Fig. 3. distribution of SAs for each of the six categories of rumors.**

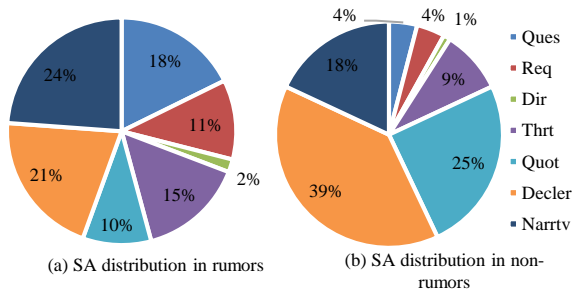

(a) SA distribution in rumors  (b) SA distribution in non-rumors

**Fig. 4. A demonstration of the distribution of speech acts in rumors (a) and non-rumors (b).**

Figure 3 shows the distribution of seven SA classes of rumors in different categories. Also, Fig. 4 shows the average distribution of SA classes in rumors (4-a) and non-rumors (4-b). As can be seen from Fig. 4-a, Persian rumors are often expressed by three SA classes including narrative, declaration, question, threat, and in some cases with the request SA. The reason is that individuals spread rumors when they feel anxiety or threat. Also, to increase the speed of the release of a rumor, the rumor publisher asks audiences to notify the message as soon as possible to his or her relatives. Rumormongers warns the audience that if they do not inform others, a bad event is may happen. Or, they may distort the audience's mind by expressing deviant questions about the issue.

### D. Student's t-test

Since our samples are independent, an independent samples T-test is run on the proposed features. An Independent Samples t-test compares the means and P-value of each feature for two groups rumor and non-rumor. Student's paired *t*-test is performed on seven classes of SA. NULL hypothesis is rejected if $P < 0.01$. In this study, the null hypothesis is defined as follows:

*Null hypothesis*: speech act is the same in rumors and non-rumors.

Therefore, Student's t-test is carried out and the result of P-value for each of the speech acts listed in Table V, with the hypothesis that each SA appears with a different frequency in FRs and TRs and can discriminate between them. The P-value results (<=0.05) demonstrate that most of these speech acts reveal statistically significant differences between FR and TR texts.

TABLE V. AVERAGE OF FREQUENCY OF SEVEN CLASSES, ALONG WITH THE P-VALUE OF A STUDENT T-TEST FOR COMPARING FREQUENCIES IN FALSE RUMOR (FR) VERSUS TRUE RUMOR (TR) GROUPS

|  | SA-Ques | SA-Req | SA-Dir | SA-Thre | SA-Quot | SA-Dec | SA-Narrtv |
|---|---|---|---|---|---|---|---|
| Average frequency in FR | 0.070 | 0.002 | 0.008 | 0.038 | 0.038 | 0.318 | 0.545 |
| Average frequency in TR | 0.024 | 3.27E-05 | 0.002 | 0.014 | 0.040 | 0.572 | 0.348 |
| P-value | 2.53E-05 | 0.159 | 0.083 | 0.003 | 0.433 | 9.92E-21 | 2.30E-13 |

According to the results of the T-test (p-value = 0.05) on seven SA classes that are listed in Table V; question, threat declaration, and narrative SAs are selected as features that can distinguish rumors and non-rumors. Therefore, the null hypothesis is rejected. This means that these four SA classes can be used in the process of rumor detection.

### E. Application of the Speech Act in Identifying Rumors

The study also aims to demonstrate the application of the speech act as a feature pragmatic to identify rumors. Based on the obtained results in the previous section, rumors are often expressed in three SA classes, including declaration, question, threat, and in some cases with the request SA. Rumor texts may have more than one SA, so SA of the text T as a double value (in range 0 to 1) is calculated in these four SA classes.

Two types of experiment are carried out to assess the effect of four SA classes in the process of rumors identification. In the first experiment, the classification of two classes of rumor and non-rumor is performed based on common context features including, negative and positive sentiment, negation (It may be present in all units of language, e.g., words (e.g., not, no, never, incredible), affixes (e.g., -n't, un-, any-), uncertainty and certainty-related words, lexical diversity (the percentage of unique words or terms in all words or terms), pronoun, depth of dependency tree, word and Sentence length, punctuation, number of words and sentences, adjective, adverb, and verb.

In the second experiment, a combination of common context features along with four SA classes are utilized to classify rumors and non-rumors. Table VI demonstrates the result of the evaluated metrics of Precision, Recall, and F-measure by RF classifier to evaluate four SA classes and their impact on identifying rumors.

TABLE VI. DIFFERENT IMPACT OF FEATURES TO DETECT TELEGRAM RUMORS BY RF CLASSIFIER.

|  | Precision | Recall | F-Measure |
|---|---|---|---|
| **(1) Common context features** | | | |
| Rumor | 0.766 | 0.753 | 0.759 |
| Non-rumor | 0.757 | 0.770 | 0.764 |
| Avg. | 0.762 | 0.762 | 0.762 |
| **(2) Common context features + Four SA classes** | | | |
| Rumor | 0.810 | 0.760 | 0.784 |
| Non-rumor | 0.774 | 0.823 | 0.798 |
| Avg. | 0.792 | 0.791 | **0.791** |

As shown in Table VI, the four SA features had a significant effect on improving the results of the classification of rumor and non-rumor. Therefore, the classification accuracy using the combination of common context features and four SA classes as new features is improved from 0.762 to 0.791 using RF classifier. This result demonstrates the positive impact of new proposed features on improving the classification accuracy.

## VII. CONCLUSION

In this study, the problem of SA classification is investigated in the Persian language in seven classes. For this purpose, a content-based method is proposed to detect rumors. In this model, a dictionary of content features is created. The WordNet ontology is used to enrich this dictionary by extracting synonymous words and inserting in the dictionary.

Four types of experiments have been performed to illustrate (1) the impact of WordNet ontology on the results of the classification of rumors and non-rumors, (2) the performance of proposed SA classifier compared to the Soltani Panah et al. [1] classifier, (3) the usual SAs in rumors based on the results of experimental analysis and T-test, (4) the impact of these SAs as effective features on the distinction between rumors and non-rumors, and its application in the identification of rumors and non-rumors.

In the first experiment, in the fourth experiment, two different evaluations (with and without WordNet) showed the positive effect of WordNet ontology on the categorization results.

In the second experiment, to improve the performance of SA classifier, we utilized the feature extraction methods to extract effective features and also used WordNet to extract synonym words to enrich the features dictionary. Experimental results prove that the FS methods as the basis for text representation and WordNet as a lexical ontology to extract the synonyms of each word within text, as well as RF and SVM as the best classifiers yielded an accuracy improvement of 0.95 in compared to presented work by Soltani-Panah[1] for SA classification on the Persian language.

In the third experiment, the proposed FS_WN_SA classifier is applied to determine the common SAs in Persian rumors. For this purpose, we first examined rumors in political, economic, events, sports, cultural, and health and medicine categories. Then, using the FS_WN_SA classifier, we identified the SAs of each category. Based on the results of classification in the seven SA classes, it was found that rumors are often expressed in three SA classes, including narrative, question, and threat, and in some cases, with the request SA. Since there is no major difference in expressing the declarative and narrative SAs, so rumor texts are expressed with a relatively similar percentage in these two SA classes. Non-rumors texts are also expressed in declarative SA. On the other hand, since we intend to use discriminating SAs between rumors and non-rumors to identify rumors, so we did not consider the declarative as a common SA in rumors. Also, the results of T-Test show that SA type of a text has the significant distinction between the means of two groups of rumor and non-rumor in four classes including, narrative, question, threat, and request.

The fourth experiment shows that these SAs can be useful as new factors to identify rumors. Thereby, this study utilized the combination of common context features and four classes of SAs to identify rumors. The experimental results are demonstrated the acceptable effect of these features on rumor detection.

As future work, we will use this SA classifier as the basis for rumors verification in the Persian language. Since, the training data plays an important role in learning a model. Training data must be labeled and large enough to cover all the upcoming classes. So, another future work is to use semi-supervised methods for SA classification.

# یک دسته‌بند کنش گفتار برای متون فارسی و کاربرد آن در شناسایی شایعات

زلیخا جهانبخش نقده[1]، محمدرضا فیضی درخشی[2*]، آرش شریفی[3]

۱- گروه مهندسی کامپیوتر، واحد علوم و تحقیقات تهران، دانشگاه آزاد اسلامی، تهران، ایران.

۲*- گروه مهندسی کامپیوتر، دانشکده مهندسی برق و کامپیوتر، دانشگاه تبریز، تبریز، ایران.

۳- گروه مهندسی کامپیوتر، واحد علوم و تحقیقات تهران، دانشگاه آزاد اسلامی، تهران، ایران.

[1]zoleikha.jahanbakhsh@srbiau.ac.ir, [2*] mfeizi@tabrizu.ac.ir, and [3]a.sharifi@srbiau.ac.ir

* نشانی نویسنده مسئول: محمدرضا فیضی درخشی ، تبریز، بلوار ۲۹ بهمن، دانشگاه تبریز، دانشکده مهندسی برق و کامپیوتر.

**چکیده**– کنش گفتار یکی از حوزه های مهم منظور شناسی زبان است که به ما درک درستی از وضعیت ذهن فرد و انتقال عمل زبانی مورد نظر می دهد. آگاهی از کنش گفتار یک متن می تواند در تجزیه و تحلیل آن متن در برنامه های کاربردی پردازش زبان طبیعی مفید باشد. این مطالعه یک روش آماری مبتنی بر دیکشنری برای شناسایی کنش‌های گفتاری در متون فارسی ارائه می دهد. در روش پیشنهادی، کنش گفتاری یک متن بر اساس چهار معیار شامل، ویژگی های لغوی، نحوی، معنایی و سطحی و با استفاده از تکنیک‌های یادگیری ماشین، در هفت کلاس کنش گفتار مورد ارزیابی قرار می‌گیرد. همچنین، از آنتولوژی ووردنت برای غنی‌سازی دیکشنری ویژگی‌ها استفاده می‌شود. به این صورت که، مترادف کلماتی که در دیکشنری ویژگی‌ها وجود ندارند استخراج شده و با لغات موجود در دیکشنری تطبیق داده می‌شود. برای ارزیابی تکنیک پیشنهادی، از چهار روش دسته‌بندی شامل جنگل تصادفی (RF)، ماشین بردار پشتیبان(SVM)، نایو بیز(NB) و K نزدیک ترین همسایه (KNN) استفاده شده است. نتایج تجربی نشان می دهد که روش پیشنهادی با استفاده از RF و SVM به عنوان بهترین دسته‌بندها، عملکرد پیشرفته ای با میانگین F-measure ۰.۹۵ برای دسته‌بندی متون فارسی بر اساس کنش گفتار دارد. دیدگاه اصلی ما از این کار، معرفی یکی از کاربردهای شناسایی کنش گفتار در محتوای رسانه های اجتماعی، به ویژه کنش گفتار رایج در شایعات و کاربرد آن در تشخیص شایعات است. نتایج نشان داد که شایعات فارسی اغلب با سه کلاس کنش گفتار روایتی، سوالی و تهدیدی و در برخی موارد با کنش گفتار درخواستی بیان می شوند. همچنین نتایج ارزیابی نشان می دهد که SA به عنوان یک ویژگی متمایزکننده بین شایعات و غیر شایعات ، صحت شناسایی شایعه را از ۰.۷۶۲ (بر اساس ویژگی های متنی رایج) به ۰.۷۹۱ (ترکیبی از ویژگی های رایج و چهار کلاس SA ) بهبود می بخشد.

**واژه‌های کلیدی**: کنش گفتار، دسته‌بندی متون فارسی، استخراج ویژگی، ووردنت، تشخیص شایعه.